\documentclass[runningheads]{llncs}
\usepackage{url}
\usepackage{cite}
\usepackage{amsmath,amssymb,amsfonts}
\usepackage{graphicx}

\usepackage{subcaption}
\usepackage{textcomp}
\usepackage{multirow}

\usepackage{algorithm}
\usepackage{algpseudocode}

\usepackage{amsmath}
\usepackage{xcolor}
\usepackage[misc]{ifsym}

\begin{document}
\title{Noise-NeRF: Hide Information in Neural Radiance Field using Trainable Noise}
%
%
\author{Qinglong Huang\inst{1,2} \and
Haoran Li\inst{1,2} \and
Yong Liao\inst{1,2,{(\textrm{\Letter})}} \and
Yanbin Hao\inst{1} \and
Pengyuan Zhou\inst{3,{(\textrm{\Letter})}}
}
\institute{
University of Science and Technology of China, Hefei, China \\
\email{\{qinglonghuang,lhr123,\}@mail.ustc.edu.cn, yliao@ustc.edu.cn, haoyanbin@hotmail.com} \and
CCCD Key Lab of Ministry of Culture and Tourism, Hefei, China \and
Aarhus University, Aarhus, Denmark \\
\email{pengyuan.zhou@ece.au.dk} }

\maketitle              
\begin{abstract}
Neural Radiance Field (NeRF) has been proposed as an innovative advancement in 3D reconstruction techniques. However, little research has been conducted on the 
issues of information confidentiality and security to NeRF, such as steganography. Existing NeRF steganography solutions have shortcomings in low steganography quality, model weight damage, and limited amount of steganographic information. This paper proposes Noise-NeRF, a novel NeRF steganography method employing Adaptive Pixel Selection strategy and Pixel Perturbation strategy to improve the quality and efficiency of steganography via trainable noise. Extensive experiments validate the state-of-the-art performances of Noise-NeRF on both steganography quality and rendering quality, as well as effectiveness in super-resolution image steganography.

\end{abstract}
\section{Introduction}
\label{sec:intro}
The neural radiance field (NeRF) \cite{tancik2023nerfstudio} can reconstruct three-dimensional photo-realistic scenes from limited 2D images taken from different viewpoints with scene continuity \cite{barron2022mip}. NeRF holds great potential in digital media such as virtual reality, augmented reality, special effects games, etc \cite{levy2023seathru}.

Meanwhile, the information confidentiality and data security issues of NeRF have garnered increasing attentions~\cite{horvath2023targeted}. NeRF steganography is one of such challenges and has seen few studies only from recently~\cite{luo2023copyrnerf,li2023steganerf}. Current approaches based on retraining the NeRF model have \textbf{three shortcomings}: 1) their embedded information into the model weights inevitably damage the model, resulting in unstable reconstruction qualities under different viewing angles \cite{li2023steganerf}; 2) they can hide limited amount of steganographic information. Current methods mainly embed information in a single image or binary code for a single NeRF scene, which would face quality collapse when embedding too much information; 3) they mainly work with low-quality images while hiding information in super-resolution images is still overlooked.

\begin{figure}[th!]
\vspace{-1.5em}
\centerline{\includegraphics[width=0.9\textwidth]{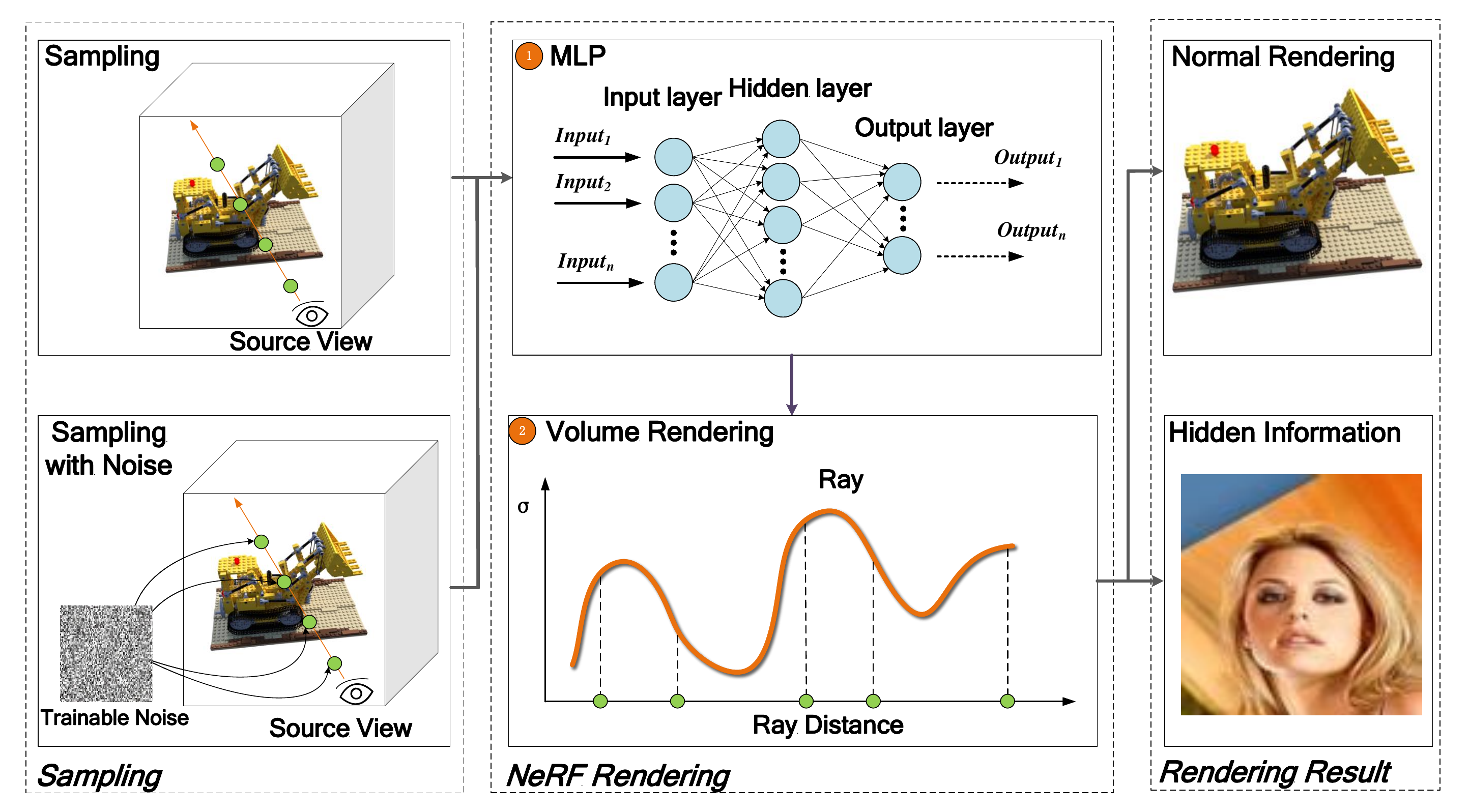}}
\caption{Overview of Noise-NeRF.}
\label{fig1}
\vspace{-1.5 em}
\end{figure}

To deal with the mentioned challenges, this paper proposes a novel NeRF steganography method namely Noise-NeRF based on trainable noise, as shown in Fig.~\ref{fig1}. Noise-NeRF takes advantage of the neural networks in NeRF to query color and density information. We introduce trainable noise on specific views to achieve information steganography. Specifically, the NeRF model renders the secret information when we input the noise during sampling, otherwise renders the normal images. Noise-NeRF only requires to update the input noise without changing any weight, thus does not impact the rendering quality. To address the varying sensitivity of different pixels to embedded noise, we propose an Adaptive Pixel Selection strategy to ensure the steganography accuracy. Furthermore, we introduce a Pixel Perturbation strategy to accelerate the convergence with trainable noise. Our contributions  can be summarized as follows:
\begin{itemize}
\item We propose the first lossless NeRF steganography method namely Noise-NeRF, by updating the input noise at a specific view instead of changing the model weights like other proposals. Our method ensures the NeRF model achieves information steganography without impacting its rendering quality.
\item We propose an Adaptive Pixel Selection strategy and a Pixel Perturbation strategy to select pixels with greater differences according to the gradient to update the noise. We update the input noise in the early stage and finely process the pixel details of hidden content in the later stage. Our strategies significantly improve the recovery quality and steganography efficiency of NeRF.
\item We conduct extensive experiments on ImageNet and several famous super-resolution image datasets using a series of pre-trained NeRF scenes. The results demonstrate the superior performance of Noise-NeRF in both steganography quality and rendering quality.
\end{itemize}
\section{Related Work}
\subsection{Neural Radiance Field}
The success of NeRF\cite{mildenhall2021nerf} has drawn widespread attention to the simple and high-fidelity three-dimensional reconstruction method of neural implicit representation. Implicit representation is a continuous representation that can be used for the generation of new perspectives and usually does not require 3D signals for supervision. NeRF realizes an effective combination of neural fields and graphics component volume rendering\cite{mildenhall2021nerf}. It uses a neural network to implicitly simulate the scene. By inputting the spatial coordinates of the three-dimensional object, NeRF outputs the corresponding geometric information. There are currently many improvements and application research on NeRF, including training acceleration \cite{Chen2022ECCVtensorf,fridovich2022plenoxels,muller2022instant}, content edition \cite{bao2023sine,zhan2022general,hyung2023local}, generalization \cite{ yu2021pixelnerf,wang2021ibrnet,wang2023rodin,irshad2023neo}, and large-scale scenes \cite{mi2023switchnerf,tancik2022block}, etc. These studies have enabled efficient three-dimensional reconstruction and practical applications of NeRF in many usecases. Meanwhile, with the launch of NeRF-related products such as Luma AI\cite{luma}, issues such as information security and copyright protection for NeRF have become increasingly important.

\subsection{Steganography for 2D Image}
Steganography for 2D images is an important direction in the field of information security. Traditional image steganography methods generally use redundant information in the image to hide secret information \cite{marvel1999spread}. For example, the most popular technique is ``least significant bit'' (LSB) steganography\cite{zhang2003reliable,luo2010edge,rustad2022inverted}, which embeds secret information into the least significant bits of the pixel values of 2D images. LSB can hide a large amount of content via small changes to the image, and is difficult to detect. With the development of deep neural networks, there are also many studies using neural networks for information hiding\cite{zhu2018hidden,baluja2017hiding,baluja2019hiding}. DeepStega\cite{baluja2019hiding} can hide the steganographic image in a carrier of the same size. Nowdays, as the representation model of 3D scenes based on neural radiation fields has received widespread attention, steganography for NeRF is becoming an important research direction.

\subsection{Steganography in NeRF}
In the past, 3D scenes was mainly represented by explicit representation, such as mesh, point cloud, voxel, and volume \cite{riegler2017octnet}. These representations enable explicit modeling of scenes. They are also convenient for extending the steganography method of 2D images to 3D scenes, such as\cite{praun1999robust,zhu2021gaussian,wu2021embedding}. However, NeRF as an implicit representation functions in a completely different way. It maps the coordinate information of each point in the spatial scene to the color and density of the point. The internal weights make it difficult to accurately express the physical meaning with clear interpretability. Therefore, explicit translation, rotation, scaling, embedding, and other steganographic measures are difficult to apply to NeRF. 

StegaNeRF\cite{li2023steganerf} is the first study on hiding information in NeRF. They hide natural images in 3D scene representations by retraining NeRF parameters, and simultaneously train a decoder that can accurately extract hidden information from NeRF-rendered 2D images. In 2023, CopyRNeRF\cite{luo2023copyrnerf} studied copyright for NeRF, a research question that shares lots of similarities with steganography in NeRF. They proposed to protect the copyright of a NeRF model by replacing the original color representation with the color representation of watermarks. They use a decoder to recover the binary secret information from the rendered image while maintaining high rendering quality and allowing watermark extraction. Although effectively performing NeRF steganography, their method faces issues such as model retraining, limited amount of hidden data, and steganography quality. In this work, we propose Noise-NeRF to tackle these challenges.

\section{Method}

\subsection{NeRF Preliminary}\label{AA}
NeRF represents a continuous scene in space as a 5D neural radiation fields. It inputs the position information $(x,y,z)$ and direction information $(\psi,\phi)$ of a specific point in the scene and outputs color information $c$ and voxel density information$\ sigma$. The neural radiation fields $F_\theta$ with trainable parameters $\theta$ can be expressed as:
\begin{equation}
F_\theta: (x,y,z,\psi,\phi) \to (c,\sigma).
\label{eq1}
\end{equation}

Next, NeRF uses the volume rendering formula to sample the rays along the observation direction and passes the sampled 3D points through the neural network to obtain the pixel value $c$ and voxel density $\sigma$ of each point for sampling and superposition to finally obtain the pixel value corresponding to this ray direction:
\begin{equation}
\begin{aligned}
\hat{C}(\mathbf{r})=\sum_{i=1}^N T_i\left(1-\exp \left(-\sigma_i \delta_i\right)\right) \mathbf{c}_i, \text { where } T_i=\exp \left(-\sum_{j=1}^{i-1} \sigma_j \delta_j\right),
\label{eq2}
\end{aligned}
\end{equation}
where $\hat{C}$ denotes the color rendered by the camera ray $r(t)=o+td$, $N$ represents the number of points sampled on the ray. $\sigma_i$ represents the distance between adjacent sampling points $i$ and $i+1$.

NeRF also adopts a hierarchical sampling strategy to train and optimize the network parameters $\theta$ through the mean square error (MSE loss) between the rendered and the true pixel colors. This enables NeRF to learn implicit representations and capture the features of 3D scenes:
\begin{equation}
\begin{aligned}
L=\sum_{r \in R}\left[\left\|\hat{C}_c(r)-C_{GT}(r)\right\|_2^2+\left\|\hat{C}_f(r)-C_{GT}(r)\right\|_2^2\right],
\label{eq3}
\end{aligned}
\end{equation}
where $R$ represents all the rays in the input viewpoint, $C_c(r)$ and $C_f(r)$ represent the color prediction of the ray by the coarse network and the fine network, respectively. $C_{GT}(r)$ denotes the ground truth.

\begin{figure*}[t]
\centerline{\includegraphics[width=1\textwidth]{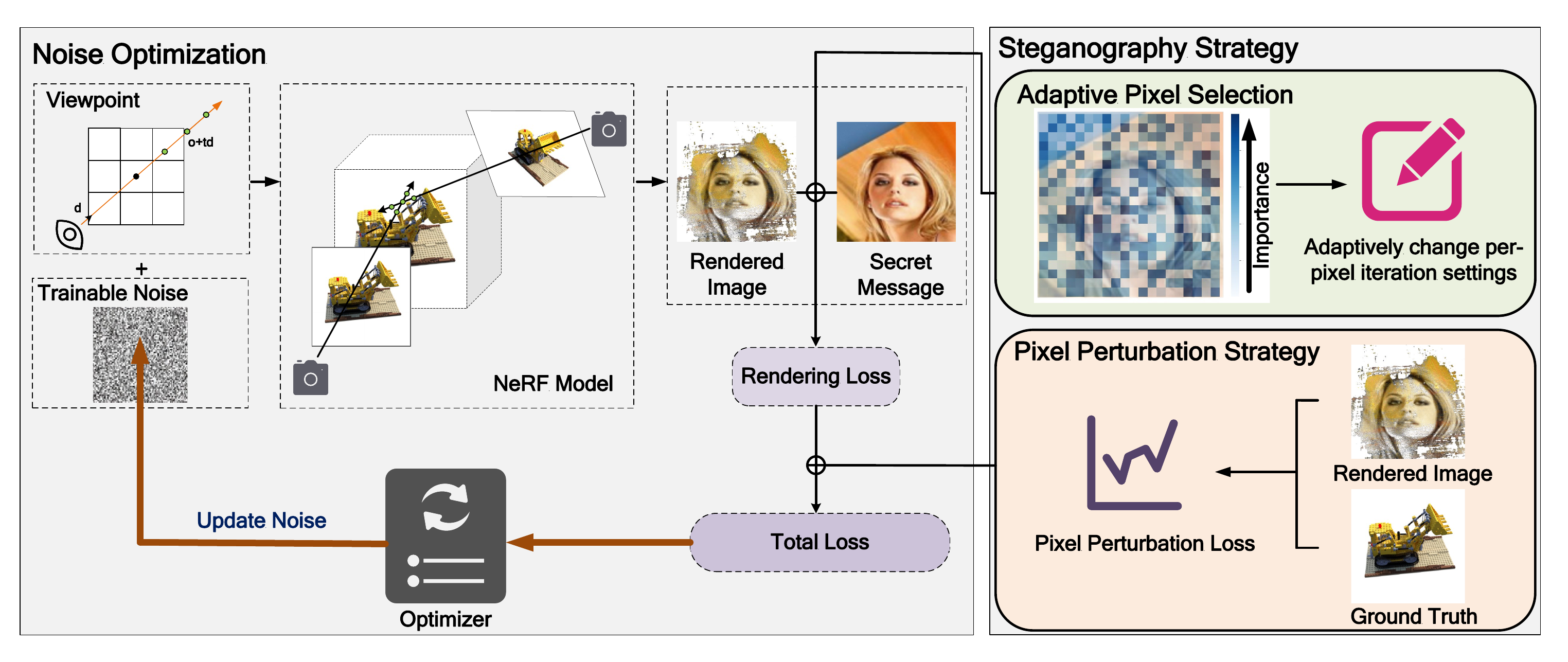}}
\caption{Framework of Noise-NeRF. We first add random initial trainable noise  to a specific view, and use pre-trained NeRF for prediction. Then, we perform supervised training on input noise using secret images. We employ Adaptive Pixel Selection strategy and Pixel Perturbation strategy during the training process to improve the quality and efficiency of steganography.}

\label{fig2}
\end{figure*}

\subsection{Noise Optimization}\label{BB}
The goal of Noise-NeRF is to embed the steganographic information into the noise by calculating the gradient and updatting the noise. Let $\theta$ denote the weight of a pre-trained NeRF scene and $M$ denote the hidden information. For a certain viewpoint $P$, a normal picture $C$ can be obtained through NeRF rendering, that is, ${f_\theta }(P) = C$. We aim to generate noise $\delta$ through Noise-NeRF so that the model can render the steganographic content $M$, that is, ${f_\theta }(P + \delta ) = M$.

The implementation framework of Noise-NeRF is shown in Fig.~\ref{fig2}. Noise-NeRF is inspired by adversarial attack\cite{zheng2019distributionally}, which is a method that makes small perturbations to the original input samples to cause the neural network to produce misclassification or wrong output\cite{madry2017towards}. We add noise under a specific viewpoint to cause the NeRF's neural network to produce intentional error output, thus NeRF can render the hidden information. Since adversarial attack examples commonly show better results in high-dimensional space\cite{carlini2017towards}, and NeRF maps low-dimensional coordinate points and directions to high-dimensional space through positional encoding\cite{mildenhall2021nerf} (Eq. \eqref{eq4}) to improve the network's ability to capture high-frequency information\cite{vaswani2017attention}, we add noise after positional encoding in Noise-NeRF.

\begin{equation}
\gamma(p)=\left(\sin \left(2^0 \pi p\right), \cos \left(2^0 \pi p\right), \cdots, \cos \left(2^{L-1} \pi p\right)\right).
\label{eq4}
\end{equation}

We add noise to the 5D coordinate points after positional encoding and then perform three-dimensional rendering through NeRF's MLP. The goal is to minimize the difference between the steganographic image and the image generated by the original NeRF by calculating the following loss.

\begin{equation}
\begin{aligned}
L_{rgb}=\sum_{r \in R}\sum_{p \in r}\left[\left\|\hat{C}_{f}(\gamma(p)+\delta)-C_{M}(\gamma(p))\right\|_2^2\right],
\label{eq5}
\end{aligned}
\end{equation}
where $R$ represents all the rays in the input viewpoint, $r$ represents one of the rays, $\delta$ is the added noise, $C_{M}$ is the steganographic target information.

Noise-NeRF calculates the gradient of the model via backpropagation to find the best direction to perturb the input sample. We then update the input noise along the direction of the gradient so that the NeRF model can produce steganographic information, as follows. 

\begin{equation}
\begin{aligned}
\delta^{i}_{p}= \delta^{i-1}_{p}+\eta \cdot({\nabla _{\delta _p^t}}{\widetilde L_{rgb}}),
\label{eq6}
\end{aligned}
\end{equation}
where $\delta _p^t$ represents the noise added to the $p$ sampling point in the $i$ iteration process, and $\eta$ is the learning rate.




\subsection{Adaptive Pixel Selection}\label{CC}
Though we calculate the gradient information of the input noise and update it through backpropagation, not all pixels are equally sensitive to the input noise. Different pixels between the steganographic target and NeRF's predicted image would cause different loss values and require different iteration settings to generate better noise. Therefore, we refer to the idea of batch size adaptation\cite{wang2020towards} and propose Adaptive Pixel Selection strategy, which adaptively selects pixels and sets different iterations.

Given a set of pixel batch sizes $S = {s_1,...,s_m}$, we select each batch size $s_i$ ($\forall {s_i} \in S$) in one iteration, compute the gradient, and update the input noise. To measure the impact of different batch sizes on steganography performance, we assume that the convergence speed remains stable within an iteration. If the batch size $s_i (s_i \in S)$ reduces the average loss the most in each query, it is considered the most appropriate batch size. Our method shares the gradients computed in the maximum batch size.

\subsection{Pixel Perturbation Strategy}\label{DD}
When updating noise, we aim to recover the steganographic information $M$ from the camera pose $P$ of the selected viewpoint. For the relatively NeRF network, using iterative loss calculation (Eq. \eqref{eq5}) and backpropagation is computationally heavy. Therefore, we target a fast deviation of the rendered image from the original image in the early stage of the noise update process. To achieve that, we need the noise to cause false positives in rendering $f_\theta(P+\delta)$ as much as possible. Therefore, we refer to the idea of batch size adaptation\cite{wang2020towards} and propose the Pixel Perturbation strategy as follows.

\begin{equation}
L_{perturb}= -\sum_{r \in R}\sum_{p \in r}\left[\left\|\hat{C}_{f}(\gamma(p)+\delta)-\hat{C}_{f}(\gamma(p))\right\|_2^2\right]
\label{eq7}
\end{equation}

As such, we increase the efficiency of steganography by combining the fast deviation of the image in the early image thanks to the Pixel Perturbation strategy, and, optimize the rendered image thanks to the Adaptive Pixel Selection strategy. The overall training loss of Noise-NeRF can be expressed as:

\begin{equation}
\begin{aligned}
\left\{ \begin{array}{l}
L = {\lambda _1} \cdot {L_{rgb}} + {\lambda _2} \cdot {L_{perturb}},\ iteration \le \mu\\
L = {L_{rgb}},\ iteration > \mu
\end{array} \right.
\label{eq8}
\end{aligned}
\end{equation}
where $\lambda_1$ and $\lambda_2$ control the weights of the two loss functions, and $\mu$ is the boundary value of iteration.

In summary, the input noise is updated through backpropagation by calculating its loss gradient. This can generate the noise that causes the neural network to output incorrectly, and achieve lossless steganography in NeRF. Further, we propose Adaptive Pixel Selection and Pixel Perturbation strategies to significantly improve the quality and efficiency of NeRF steganography. The overall process of the Noise-NeRF is summarized in Algorithm \ref{alg}.

\begin{algorithm}
\caption{Noise-NeRF on a single scene} 
\label{alg}
\begin{algorithmic}
\State \textbf{Input:} Pretrained NeRF model $f$ and weights $\theta$, Secret Message $M$, Viewpoint $P$
\State \textbf{Output:} Well-trained noise $\delta$
\For{each iteration $t$}
\State Conduct Adaptive Pixel Selection
\State Add noise to NeRF rendering $f_{\theta}(P+\delta_{p})$
\State Compute rgb loss $L_{rgb}$ in Eq. \eqref{eq5}
\State Compute Perturbation loss $L_{perturb}$ in Eq. \eqref{eq7}
\State Compute total loss $L$ in Eq. \eqref{eq8}
\State Update Noise $\delta^{i}= \delta^{i-1}+\eta \cdot Adam({\nabla _{\delta^t}}{L})$
\EndFor
\end{algorithmic}
\end{algorithm}

\section{Experiments}
\subsection{Implementation Details}
\noindent\textbf{Datasets and Hyperparameters.}
We chose the standard NeRF as the experimental object. For forward and 360° scenes, we selected scenes in LLFF\cite{mildenhall2019local} and NeRF-Synthetic\cite{mildenhall2021nerf} as objects respectively. We randomly selected images from imagenet\cite{deng2009imagenet} as steganographic targets. We also selected several popular super-resolution datasets: DIV2K\cite{agustsson2017ntire}, OST\cite{wang2018recovering}, FFHQ\cite{karras2019style}, CeleA-HQ\cite{karras2017progressive} to test the super-resolution steganography performance of Noise-NeRF. 
The hyperparameters in Eq.\eqref{eq8} are set as $\lambda_1$ = 0.5, $\lambda_2$ = 0.5, and $\mu$ = 50. We use the Adam optimizer, the learning rate of each iteration is set to 1e-2, and the learning decay rate is set to 0.3. All the experiments were conducted on a server equipped with an NVIDIA RTX3090 GPU.

\noindent\textbf{Metrics.}
We use PSNR, SSIM\cite{Wang_Bovik_Sheikh_Simoncelli_2004}, and LPIPS\cite{Zhang_Isola_Efros_Shechtman_Wang_2018}, the classic indicators for measuring 3D reconstruction quality in NeRF, to evaluate the NeRF rendering effect. We use SSIM and SNR to evaluate the recovery quality of steganographic information. 

\noindent\textbf{Baselines.}
For the current SOTA method StegaNeRF\cite{li2023steganerf}, we use its original settings; for the traditional algorithm LSB\cite{fridrich2001detecting} for two-dimensional pictures and the deep learning algorithm DeepStega\cite{baluja2017hiding}, we hide the information in the two-dimensional images of the training dataset, and then use the traditional NeRF training method. 

\subsection{Multiple Scenes Steganography}

\begin{figure*}[htbp]
\centerline{\includegraphics[width=1\textwidth]{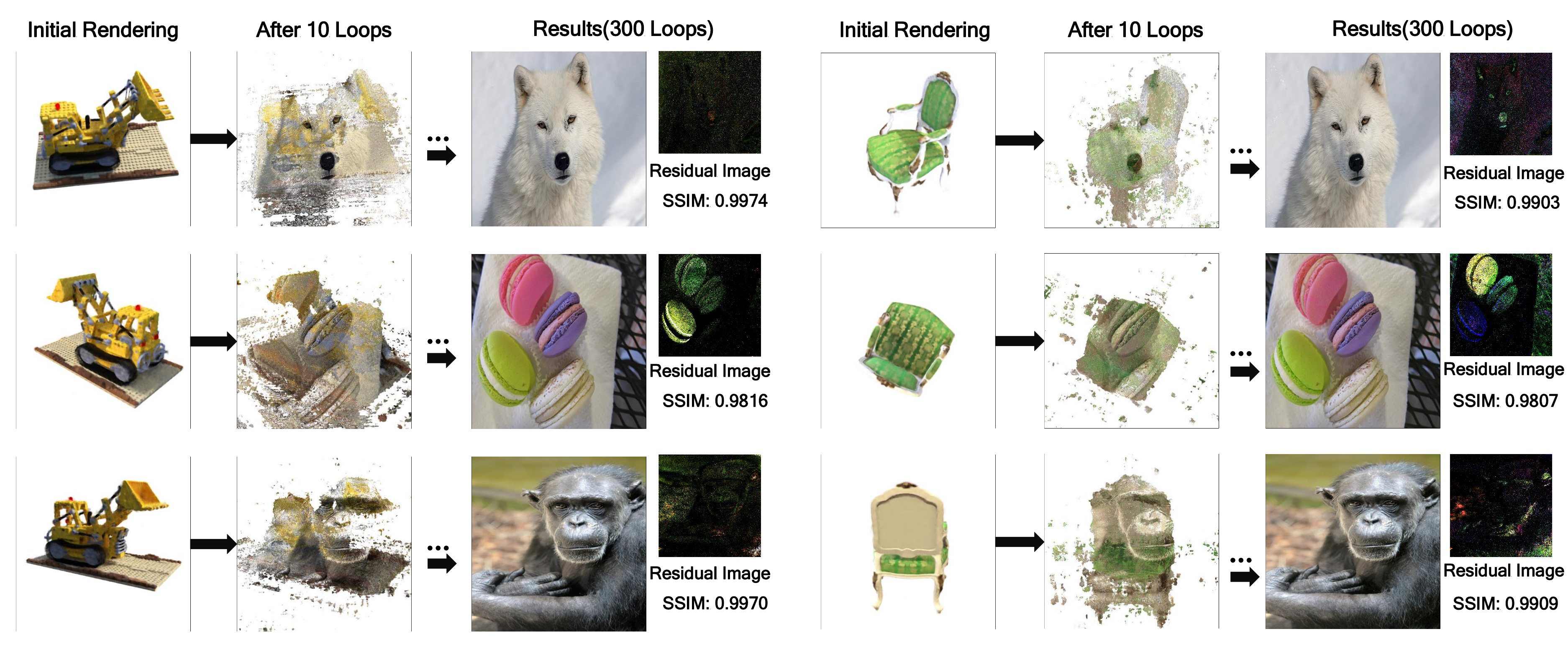}}
\caption{Noise-NeRF performances on multiple scenes. Each column displays the initial rendering, rendering after 100 loops, rendering after 300 loops, and the residual image. We also show the SSIM between the steganography image rendered by Noise-NeRF and the real hidden image.}
\label{fig4}
\end{figure*}

We first selected specified viewpoints on different scenes and used Noise-NeRF to generate noise. Then we input the noise into the NeRF model to render a steganographic image to verify steganography quality. In addition, we also test the performance of each baseline on NeRF rendering quality. The qualitative and quantitative results are shown in Fig.~\ref{fig4} and Table~\ref{tab1}.

\begin{table}[htbp]
\renewcommand{\arraystretch}{1.2}
\setlength{\tabcolsep}{6pt}
\caption{Performance comparisons on multiple scenes. Standard NeRF is an initial NeRF scenario trained with standard settings. The upper part of the table is tested on the NeRF-Synthetic dataset; the lower part is tested on the LLFF dataset. The results are the average values across different scenes.}
\begin{center}
\begin{tabular}{c|ccc|cc}
\hline
\multirow{2}{*}{Method}   & \multicolumn{3}{c|}{NeRF Rendering}                & \multicolumn{2}{c}{Embed Recovery} \\ \cline{2-6} 
                          & PSNR$\uparrow$           & SSIM$\uparrow$            & LPIPS$\downarrow$           & SSIM$\uparrow$             & ACC(\%)$\uparrow$         \\ \hline
Standard NeRF\cite{mildenhall2021nerf}             & 27.74          & 0.8353          & 0.1408          & N/A              & N/A             \\ \hline
LSB\cite{rustad2022inverted}                       & 27.72          & 0.8346          & 0.1420          & 0.0132           & N/A             \\
DeepStega\cite{baluja2019hiding}                 & 26.55          & 0.8213          & 0.1605          & 0.2098           & N/A             \\
StegaNeRF\cite{li2023steganerf}                 & 27.72          & 0.8340          & 0.1428          & 0.9730           & \textbf{100.0}           \\
\textbf{Noise-NeRF} & \textbf{27.74} & \textbf{0.8353} & \textbf{0.1408} & \textbf{0.9913}  & \textbf{100.0}  \\ \hline
Standard NeRF\cite{mildenhall2021nerf}             & 31.13          & 0.9606          & 0.0310          & N/A              & N/A             \\ \hline
LSB\cite{rustad2022inverted}                       & 31.12          & 0.9604          & 0.0310          & 0.0830           & N/A             \\
DeepStega\cite{baluja2019hiding}                 & 31.13          & 0.9606          & 0.0313          & 0.2440           & N/A             \\
StegaNeRF\cite{li2023steganerf}                 & 30.96          & 0.9583          & 0.0290          & 0.9677           & 99.72           \\
\textbf{Noise-NeRF} & \textbf{31.13} & \textbf{0.9606} & \textbf{0.0310} & \textbf{0.9847}  & \textbf{100.0}  \\ \hline
\end{tabular}
\label{tab1}
\end{center}
\end{table}

Fig.~\ref{fig4} shows that Noise-NeRF continuously optimizes noise through iterations. After inputting the noise, the image rendered by NeRF gradually approaches the target image. After 300 iterations, the SSIM of the rendered hidden image and ground truth are both greater than 98\%, meeting general steganography requirements.

As Table~\ref{tab1} shows, Noise-NeRF maintains consistent rendering quality with the standard NeRF. This is because NeRF performs standard rendering as long as no noise is input. On the other hand, all other methods require to modify NeRF's model weights to a certain extent, thus damaging the rendering quality. In terms of steganography quality, Noise-NeRF's SSIM on the two data sets got 0.9913 and 0.9847, respectively, proving its SOTA performance on NeRF steganography.

\subsection{Super-resolution Steganography}
\begin{table}[htbp]
\renewcommand{\arraystretch}{1.2}
\setlength{\tabcolsep}{2pt}
\caption{Noise-NeRF on super-resolution datasets. The amount of hidden information only depends on different trainable noises in our method. That is, by inputting different noises into the model, different hidden information can be rendered. Therefore, we use this to achieve the steganography of super-resolution images. The results are the average of NeRF-Synthetic and LLFF scenes.}
\begin{center}
\begin{tabular}{ccccccc}
\hline
\multicolumn{1}{c|}{\multirow{2}{*}{Scene}}     & \multicolumn{1}{c|}{\multirow{2}{*}{Dataset}} & \multicolumn{3}{c|}{NeRF Rendering}                                                                                                       & \multicolumn{2}{c}{Embed Recovery} \\ \cline{3-7} 
\multicolumn{1}{c|}{}                           & \multicolumn{1}{c|}{}                                  & PSNR                                        & SSIM                                         & \multicolumn{1}{c|}{LPIPS}                   & PSNR          & SSIM           \\ \hline
\multicolumn{1}{c|}{\multirow{4}{*}{NeRF-Synthetic\cite{mildenhall2021nerf}}} & \multicolumn{1}{c|}{DIV2K\cite{agustsson2017ntire}}                             & \multicolumn{1}{c|}{\multirow{4}{*}{27.74}} & \multicolumn{1}{c|}{\multirow{4}{*}{0.8353}} & \multicolumn{1}{c|}{\multirow{4}{*}{0.1408}} & 48.62
             & 0.9889         \\
\multicolumn{1}{c|}{}                           & \multicolumn{1}{c|}{OST\cite{wang2018recovering}}                               & \multicolumn{1}{c|}{}                       & \multicolumn{1}{c|}{}                        & \multicolumn{1}{c|}{}                        & 46.58             & 0.9748         \\
\multicolumn{1}{c|}{}                           & \multicolumn{1}{c|}{FFHQ\cite{karras2019style}}                              & \multicolumn{1}{c|}{}                       & \multicolumn{1}{c|}{}                        & \multicolumn{1}{c|}{}                        & 48.75           & 0.9889         \\
\multicolumn{1}{c|}{}                           & \multicolumn{1}{c|}{CelebA-HQ\cite{karras2017progressive}}                         & \multicolumn{1}{c|}{}                       & \multicolumn{1}{c|}{}                        & \multicolumn{1}{c|}{}                        & 46.80             & 0.9775         \\ \hline
\multicolumn{1}{c|}{\multirow{4}{*}{LLFF\cite{mildenhall2019local}}}      & \multicolumn{1}{c|}{DIV2K\cite{agustsson2017ntire}}                             & \multicolumn{1}{c|}{\multirow{4}{*}{31.13}} & \multicolumn{1}{c|}{\multirow{4}{*}{0.9606}} & \multicolumn{1}{c|}{\multirow{4}{*}{0.0310}} & 47.90             & 0.9814         \\
\multicolumn{1}{c|}{}                           & \multicolumn{1}{c|}{OST\cite{wang2018recovering}}                               & \multicolumn{1}{c|}{}                       & \multicolumn{1}{c|}{}                        & \multicolumn{1}{c|}{}                        & 44.91             & 0.9704         \\
\multicolumn{1}{c|}{}                           & \multicolumn{1}{c|}{FFHQ\cite{karras2019style}}                              & \multicolumn{1}{c|}{}                       & \multicolumn{1}{c|}{}                        & \multicolumn{1}{c|}{}                        & 47.59             & 0.9807         \\
\multicolumn{1}{c|}{}                           & \multicolumn{1}{c|}{CelebA-HQ\cite{karras2017progressive}}                         & \multicolumn{1}{c|}{}                       & \multicolumn{1}{c|}{}                        & \multicolumn{1}{c|}{}                        & 44.77             & 0.9799         \\ \hline
                                                &                                                        &                                             &                                              &                                              &                   &                \\
\textbf{}                                       & \textbf{}                                              & \textbf{}                                   & \textbf{}                                    & \textbf{}                                    & \textbf{}         &               
\end{tabular}
\label{tab2}
\end{center}
\end{table}

\begin{figure*}[htbp]
\centerline{\includegraphics[width=1\textwidth]{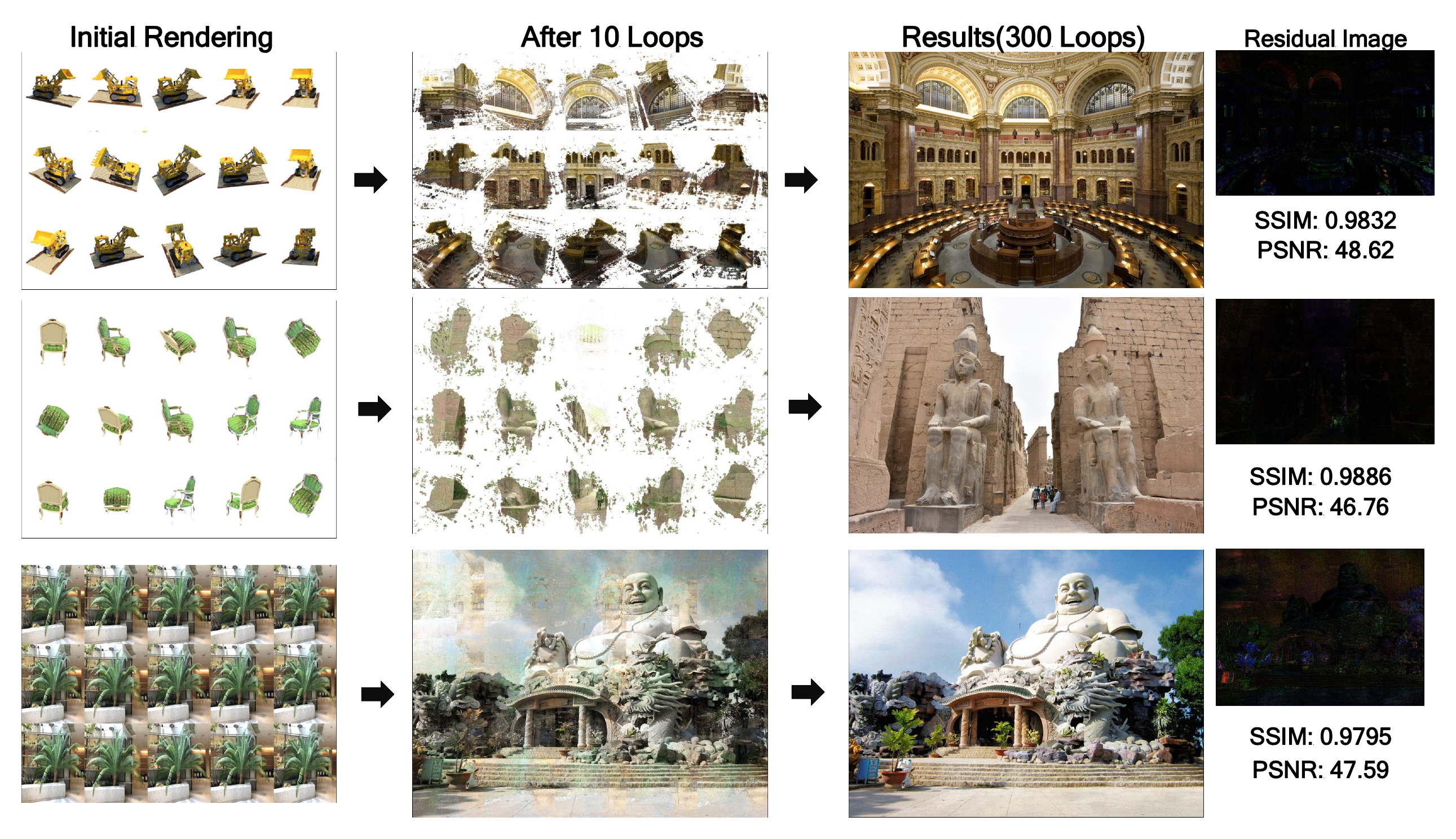}}
\caption{Noise-NeRF performance on super-resolution images. Each column displays the initial rendering, rendering after 100 loops, rendering after 300 loops, and the residual image. We also show the SSIM between the steganography image rendered by Noise-NeRF and the real hidden image.}
\label{fig5}
\end{figure*}

\begin{figure*}[htbp]
\begin{subfigure}{0.5\textwidth}
\centerline{\includegraphics[width=1\textwidth]{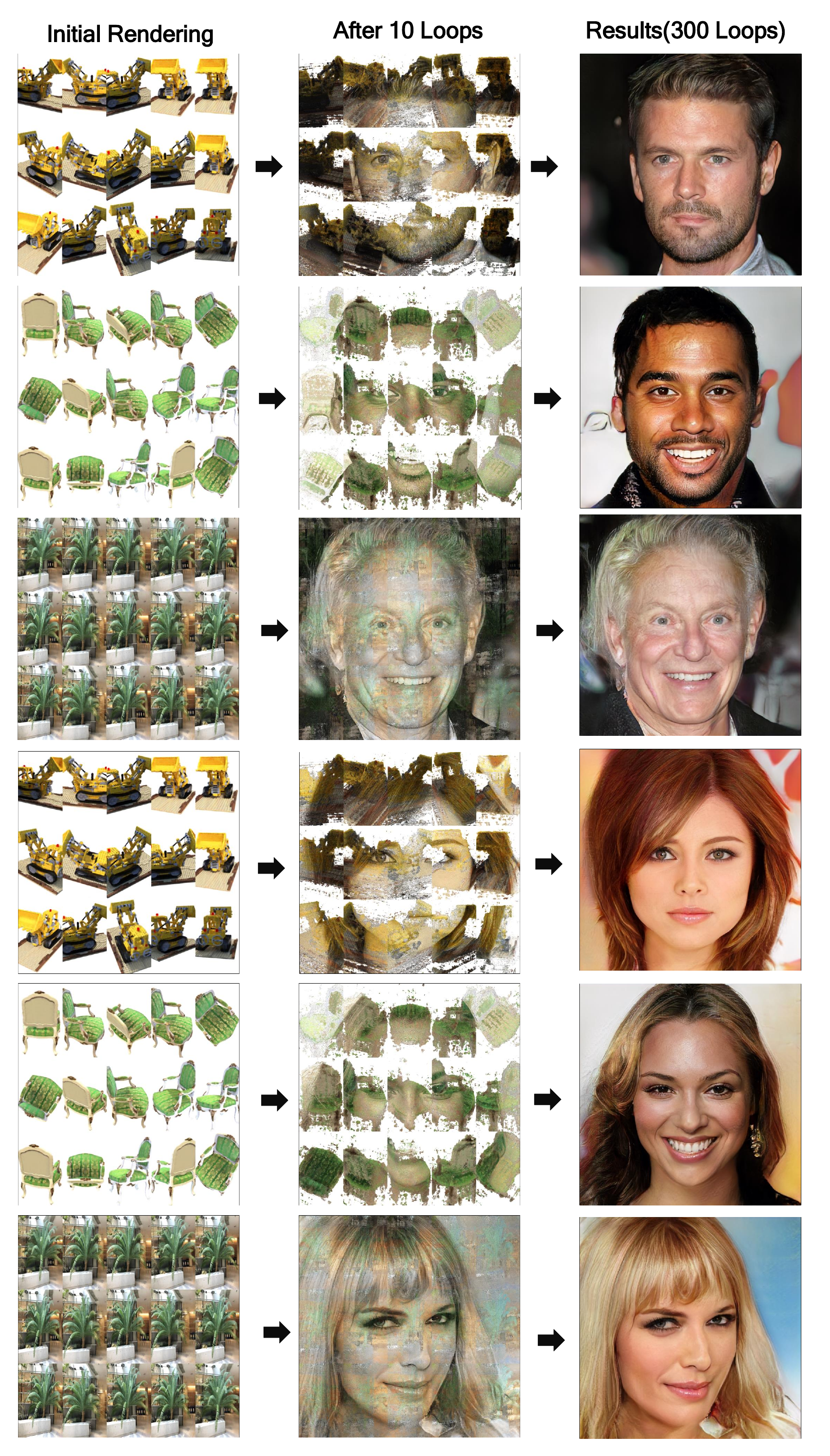}}
\end{subfigure}
\begin{subfigure}{0.5\textwidth}
\centerline{\includegraphics[width=1\textwidth]{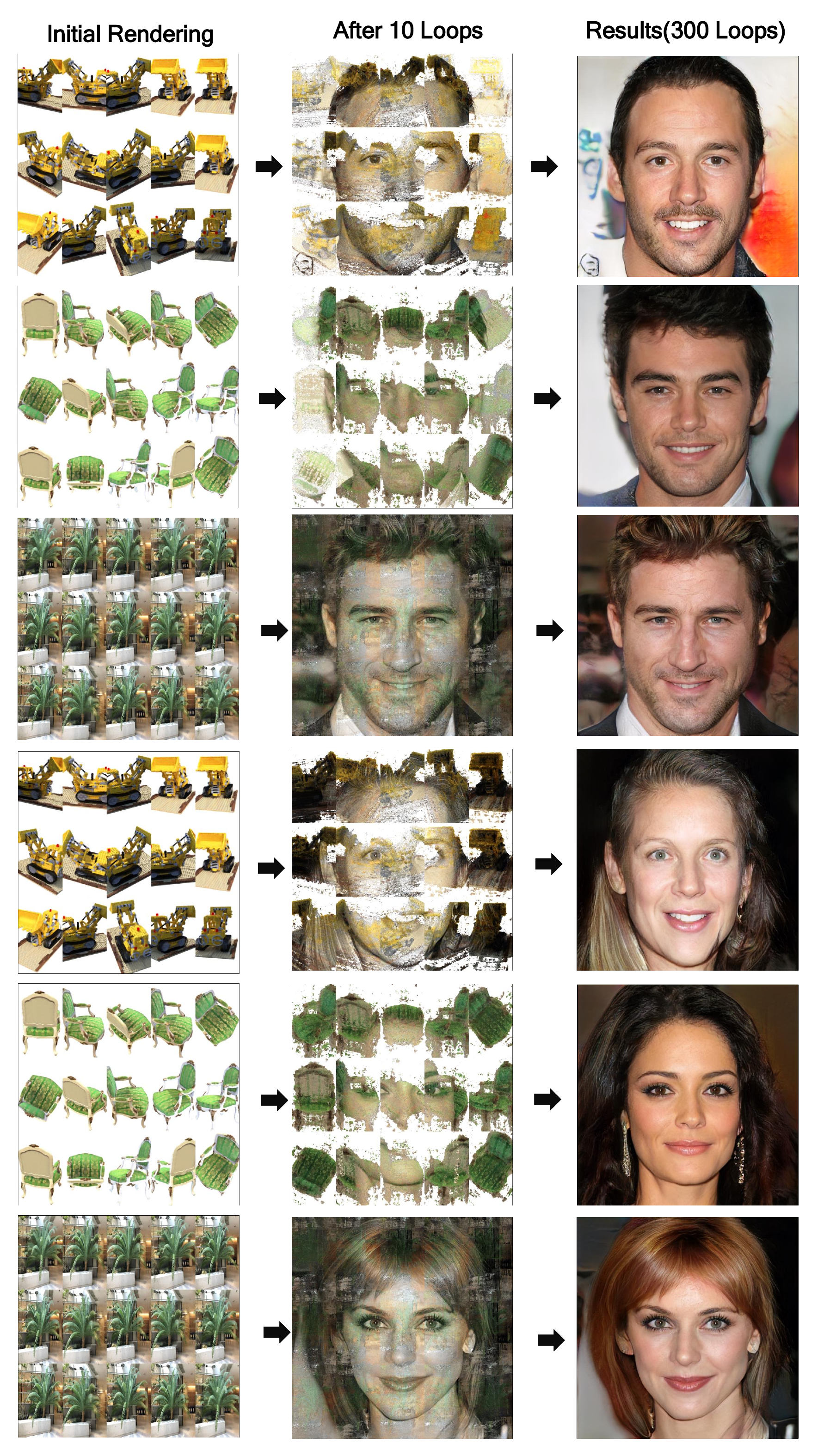}}
\end{subfigure}
\caption{More qualitative results of Noise-NeRF on multiple super-resolution results.}
\label{fig8}
\end{figure*}

In this experiment, we tested the steganography ability of Noise-NeRF on super-resolution images. We randomly selected images from the super-resolution dataset as targets, each of which had a 2K resolution. Due to the huge number of bits required for steganography, baseline steganography algorithms will make a large update to the model weights, which would cause tremendous damage to NeRF-rendered images\cite{li2023steganerf}. The visualization results of the experiment are shown in Fig.~\ref{fig5}. We clip the super-resolution image into multiple sub-images and randomly select different viewpoints of the NeRF model. We align different sub-images to different viewpoints and stitch them together to obtain the final result. As shown in Table~\ref{tab2}, in different NeRF scenes and different super-resolution datasets, Noise-NeRF achieves a 100\% success rate in NeRF steganography, with the steganographic images achieving a similarity of more than 97\%. 
Please refer to Fig.~\ref{fig8} for more details on the qualitative results. It proves the superiority of Noise-NeRF on super-resolution image steganography.


\begin{figure}[!htbp]
\centerline{\includegraphics[width=1\textwidth]{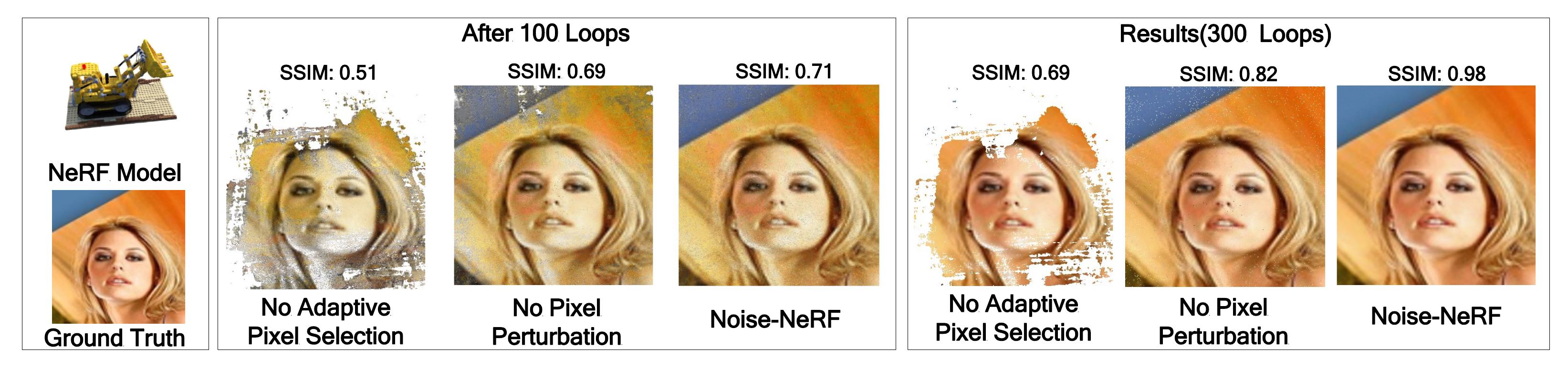}}
\caption{Ablation study of Noise-NeRF.}
\label{fig6}
\end{figure}

\subsection{Ablation Study}

We removed different components of Noise-NeRF as shown in Table~\ref{tab3} to verify the effectiveness of each part. We set the number of iterations to 300. As shown in Fig.~\ref{fig6}, we take the standard NeRF rendering image and steganographic target as a reference. From Fig.~\ref{fig6} and Table~\ref{tab3}, we observed that some pixels were completely blank in the output image without Adaptive Pixel Selection. This is because each pixel is different and has a different target pixel, thus requiring different iterations and batch size settings. Our Adaptive Pixel Selection strategy can handle this situation well by selecting pixels in a targeted manner. Removing the Pixel Perturbation strategy resulted in some pixel noise in the output image. This is because the huge neural network depth of NeRF requires many iterations of backpropagation to update the input noise and slowly gradually converge to the steganographic object. The Pixel Perturbation strategy increases the difference between the output image and the original image in the early stage, thus accelerating the noise's deviation from the original prediction of NeRF. 

\begin{table}[htbp]
\renewcommand{\arraystretch}{1.5}
\setlength{\tabcolsep}{1pt}
\caption{Ablation study of Noise-NeRF.}
\begin{center}
\begin{tabular}{c|cc|cc|cc}
\hline
\multirow{2}{*}{Method}       & \multicolumn{2}{c|}{50 Loops}  & \multicolumn{2}{c|}{200 Loops} & \multicolumn{2}{c}{Results(300 Loops)} \\ \cline{2-7} 
                              & SSIM          & Total Loss     & SSIM          & Total Loss     & SSIM               & Total Loss        \\ \hline
No strategy                   & 0.51          & 3143.79        & 0.62          & 2526.70        & 0.69               & 761.63            \\
No Adaptive Pixel Selection   & 0.44          & 976.76         & 0.49          & 211.35         & 0.59               & 83.08             \\
No Pixel Perturbation         & 0.69          & 83.67          & 0.76          & 66.15          & 0.82               & 26.40             \\ \hline
Noise-NeRF (complete version) & \textbf{0.71} & \textbf{74.33} & \textbf{0.88} & \textbf{13.24} & \textbf{0.98}      & \textbf{0.55}     \\ \hline
\end{tabular}
\label{tab3}
\end{center}
\end{table}

\section{Conclusion}
In this paper, we propose a NeRF steganography method based on trainable noise, Noise-NeRF, to address challenges faced by NeRF steganography, namely low steganographic quality, model weight damage, and limited steganographic information. We propose Adaptive Pixel Selection strategy and Pixel Perturbation strategy to improve steganography quality and efficiency. Experimental results prove the superiority of Noise-NeRF over existing baselines in both steganography quality and rendering quality, as well as effectiveness in super-resolution image steganography. 

\section{Acknowledgement}
This work is supported by the National Key Research and Development Program of China (2022YFB3105405, 2021YFC3300502).



%
%
%
\bibliographystyle{splncs04}
\bibliography{refs}

\end{document}